\documentclass[sn-mathphys,Numbered]{sn-jnl}

\usepackage{graphicx}%
\usepackage{multirow}%
\usepackage{amsmath,amssymb,amsfonts}%
\usepackage{amsthm}%
\usepackage{mathrsfs}%
\usepackage[title]{appendix}%
\usepackage{xcolor}%
\usepackage{textcomp}%
\usepackage{manyfoot}%
\usepackage{booktabs}%
\usepackage{algorithm}%
\usepackage{algorithmicx}%
\usepackage{algpseudocode}%
\usepackage{listings}%
\newcommand{\myg}{\mathcal{G}}
\newcommand{\myv}{\mathcal{V}}
\newcommand{\mye}{\mathcal{E}}

\begin{document}

\title[Representing Outcome-driven Higher-order Dependencies in Graphs of Disease Trajectories]{Representing Outcome-driven Higher-order Dependencies in Graphs of Disease Trajectories}


\author[1]{\fnm{Steven J.} \sur{Krieg}}\email{skrieg@nd.edu}

\author[1]{\fnm{Nitesh V.} \sur{Chawla}}\email{nchawla@nd.edu}

\author*[2]{\fnm{Keith} \sur{Feldman}}\email{kfeldman@cmh.edu}

\affil[1]{\orgdiv{Lucy Family Institute for Data and Society}, \orgname{University of Notre Dame}, \orgaddress{\city{Notre Dame}, \state{IN}, \postcode{46556}, \country{USA}}}

\affil[2]{\orgdiv{Children's Mercy}, \orgname{University of Missouri-Kansas City School of Medicine}, \orgaddress{\city{Kansas City}, \state{MO},  \postcode{64108}, \country{USA}}}


\abstract{The widespread application of machine learning techniques to biomedical data has produced many new insights into disease progression and improving clinical care. Inspired by the flexibility and interpretability of graphs (networks), as well as the potency of sequence models like transformers and higher-order networks (HONs), we propose a method that identifies combinations of risk factors for a given outcome and accurately encodes these higher-order relationships in a graph. Using historical data from 913,475 type 2 diabetes (T2D) patients, we found that, compared to other approaches, the proposed networks encode significantly more information about the progression of T2D toward a variety of outcomes. We additionally demonstrate how structural information from the proposed graph can be used to augment the performance of transformer-based models on predictive tasks, especially when the data are noisy. By increasing the order, or memory, of the graph, we show how the proposed method illuminates key risk factors while successfully ignoring noisy elements, which facilitates analysis that is simultaneously accurate and interpretable.}




\maketitle

\section{Introduction}
\label{sec:intro}

\begin{figure*}[ht]
    \centering
    \includegraphics[width=1.0\textwidth]{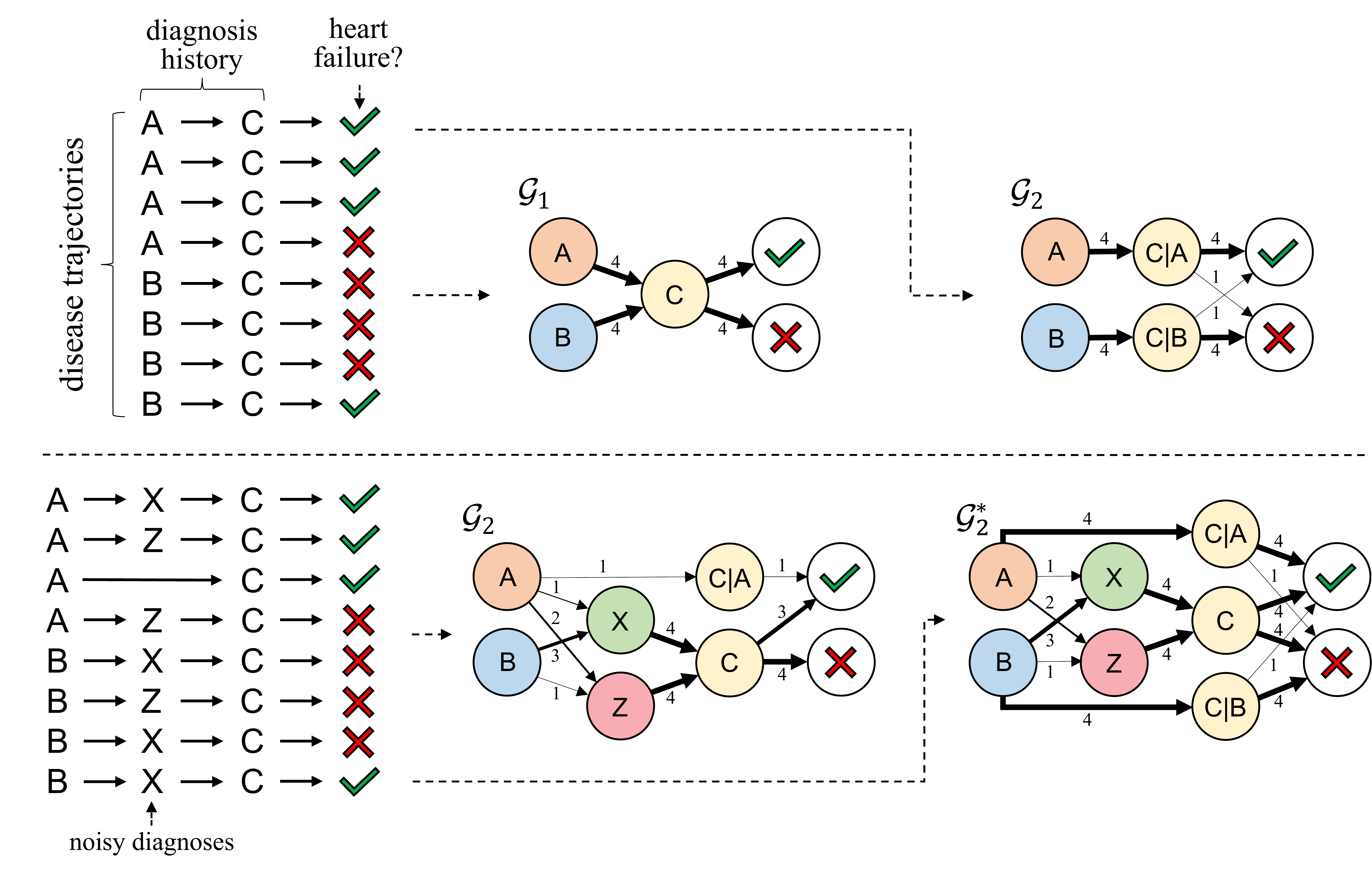}
    \caption{Given a set of disease trajectories (top), a FON $\myg_1$ cannot capture the higher-order dependency that C is more likely to lead to heart failure when it is preceded by A. A HON with $k=2$ ($\myg_2$) addresses this problem by creating the conditional nodes C$|$A and C$|$B. However, when the noisy diagnoses X and Z are present (bottom), $\myg_2$ fails to detect the importance of C$|$A and C$|$B. Our proposed model, $\myg^*_2$, addresses this problem by explicitly conditioning paths on the outcome variable and enabling them to skip noisy diagnoses. Edge weight is represented by arrow thickness.}
    \label{fig:toy}
\end{figure*}

In this era of digital medicine, computational analysis of historical patient data is a foundational approach for generating evidence-based insights into patient care, as well as developing new knowledge surrounding the etiology, risk factors, and progression of health conditions~\cite{shilo2020axes,raghupathi2014big}. While each assessment of an individual occurs at a discrete point in time, it is critical to recognize that data collected from these observations are not independent. The nature of human disease and the structure of the healthcare system itself impose temporal dependencies that connect information across an individual’s lifetime~\cite{burton2015life,henly2011health,hollar2018understanding}. As a result, appropriately utilizing historical data requires the capability to model not only the incidence of prior events but also the relationships among data over time. 

To capture these complex interactions between events over time, researchers have widely adopted supervised neural architectures~\cite{allam2021analyzing, amirahmadi2023deep}. In contrast to traditional, unsupervised trajectory models such as latent growth curves, group-based trajectory models, and temporal clustering~\cite{van2020overview,nguena2020trajectory}, these techniques are designed to directly learn relationships between patterns in sequential data and outcome incidence. Building on successes of recurrent neural network (RNN) architectures like long short-term memory (LSTM), transformer models have achieved state-of-the-art results in modeling sequence data by leveraging innovations such as self-attention and positional encodings~\cite{vaswani2017attention,devlin2018bert}. In the healthcare domain, models such as BEHRT and Med-BERT have translated the successes of BERT (Bidirectional Encoder Representations from Transformers) to tasks like disease prediction~\cite{li2020behrt,rasmy2021med,meng2021bidirectional}. While these models have demonstrated a strong ability to predict if a future event will occur, they fail to provide insight surrounding how the outcome was reached. 

The process, however, of systematically identifying these pathways across a range of health conditions presents a significant challenge. As electronic health records collect larger and more heterogeneous volumes of personalized data, temporal patterns become obfuscated by an intractable number of paths connecting information collected over time. This challenge is compounded by the fact that real-world data often contains gaps along relevant sequences, barring progression in a perfectly linear fashion, which can introduce significant amounts of noise to a trajectory. For example, diagnoses made during an emergency room visit following a car accident may interrupt sequences within a multi-year trajectory of congestive heart failure. While attention mechanisms can be used to capture relationships of data to an outcome, the specific dependencies between steps in a sequence remain difficult to extract, as the weights from the local interactions across each distinct attention head oversimplify the complex set of multilayer computations that are used to make a given prediction~\cite{chefer2021transformer}. 

Seeking more efficient and interpretable representations of sequence data, researchers have leveraged graphs (networks) to disentangle these overlapping and heterogeneous relationships. This approach has found success in many tasks such as predicting future comorbidities \cite{thomas2018predicting}, inferring causality~\cite{kannan2016conditional}, identifying disease subtypes~\cite{li2015identification}, and trajectories~\cite{jensen2014temporal,vlietstra2020identifying}. 
Despite their advantages, graphs often underfit the complex relationships that govern trajectories~\cite{capobianco2015comorbidity, hu2016network}, as they operate on a first-order Markovian paradigm, such that the next step in a sequence depends on probabilities from the current state (Fig \ref{fig:toy}). To address this, higher-order networks (HONs) were proposed to detect dependencies (i.e., conditional probabilities) across multiple steps in a sequence, and encode those dependencies in a graph~\cite{xu2016representing}. Using this approach enables far more robust and interpretable trajectories to be built in a computationally feasible manner. HONs and similar approaches have found early applications in several tasks both within and beyond the healthcare domain, including modeling diabetes comorbidities~\cite{krieg2020higher}, anomaly detection for medication order~\cite{niu2020anomaly}, and many more~\cite{rosvall2014memory,lambiotte2019networks,saebi2020efficient,krieg2022deep}. Unfortunately, HONs exhibit a fundamental limitation, as they capture transitions between steps in a strictly linear fashion. Thus, without knowing \textit{a priori} which data are relevant for a particular pathway, existing HONs cannot detect which steps are most relevant for a particular long-term outcome, and likewise cannot robustly handle the noisy probabilities that we observe in real-world health data.

Inspired by the generalizability and interpretability of graphs, this manuscript proposes a novel, supervised implementation of a HON to address exactly this. First, it introduces a method for extracting the steps along a trajectory that are most informative of an outcome, even in the presence of noisy or irrelevant subsequences. Next, it demonstrates how the interpretable path information learned by our noise-resilient model can be integrated into state-of-the-art transformer models like BERT, to further improve already strong estimates of outcome probability. Additionally, it shows that our proposed method can be used to solve more general problems, like isolating key risk factors along a trajectory of diagnoses. Ultimately, the manuscript concludes with a discussion regarding the future extensions of the proposed method for complex care patterns and translation into other clinical scenarios.\footnote{Code is publicly available at \url{https://github.com/sjkrieg/growhonsup}.}

\section{Higher Order Networks: An Overview}

To facilitate analysis, digital health data have commonly been structured to mirror paper records by characterizing an individual as a set of discrete data elements. These elements can then be mapped easily to transactional database structures for which statistical and machine-learning methods have predominantly been developed. Although possible, the ability to model conditional relationships between two (or more) events requires large and highly sparse arrays of data representing pre-defined combinations of all data elements. Consequently, network analysis has gained recognition due to its ability to model observed co-occurring data without defining the entire matrix of possibilities. To do so, networks typically utilize directed graphs. Yet these representations operate under a first-order Markovian paradigm, implying transitions to a future next step are a direct result of the current position. This limits the capacity of these networks to capture temporal trends and explicate progression. For example, Figure \ref{fig:toy} shows a typical graph $\myg_1$, or a first-order network (FON). In these disease trajectories, the transition $A \xrightarrow[]{} C$ is the key predictor of heart disease; however, because graphs rely on transitive paths of pairwise connections, the structure of $\myg_1$ cannot account for this.

To address the limitations of FONs, Xu et al.~\cite{xu2016representing} proposed higher-order networks (HONs), which create conditional nodes (e.g., $B|A$, read as $B$ given $A$) to encode higher-order dependencies (i.e., conditional probabilities in sequence data) in a graph \cite{xu2016representing,saebi2020efficient}. Intuitively, a HON is a graph with memory. The difference between a conditional node like $B|A$ and its lower-order relative $B$ is typically quantified via relative entropy over the transition probabilities of its future states~\cite{krieg2020growhon}. For example, consider a diagnosis of hypertension (ICD9 code 401). When this diagnosis is preceded by a diagnosis of renal failure (ICD9 code 584), it is more likely to be succeeded by other kidney-related diseases~\cite{krieg2020higher}. If this change in probability is sufficiently high across all diagnoses, the relative entropy would be high, and the HON would create a conditional node $401|584$. While HON models have produced new insights in many domains, they have been limited to unsupervised learning settings. When considering disease trajectories, we are often interested in studying not just divergent probabilities at each step, but \textit{differences in outcomes} across all patients. For example, if we are studying factors that lead to heart disease, even if the transition probability distribution of $401|584$ differs from $401$, there is no guarantee that this difference provides any meaningful information about the risk factors or progression of the disease. Further, although the conditional nodes in a HON allow it to consider more complex interactions between historical data, prior models a strictly linear, meaning that once data are entered into the trajectory, a HON cannot detect or skip irrelevant elements. As the HON $\myg_2$ exemplifies in Figure \ref{fig:toy}, this results in fragmented paths that often fail to capture the key relationships that drive the outcome. Constructing a HON in a supervised learning setting thus poses the following key challenges:
\begin{enumerate}
    \item How to define and detect outcome-driven higher-order dependencies (i.e., how to define the set of conditional nodes).
    \item How to encode the higher-order dependencies (i.e., how to connect the conditional nodes).
\end{enumerate}

\section{Methods}
\subsection{Preliminaries}
Let $\mathcal{S} = \{S_1, S_2, ..., S_n\}$ be a set of observed \textbf{trajectories}, where each $S_i = \langle s_1, s_2, ..., s_m \rangle $ is a sequence of \textbf{entities} such as diagnosis codes. Let $\mathcal{A} = \bigcup \mathcal{S}$ denote the set of entities across all sequences. In the supervised learning setting, we additionally consider an outcome or class variable $Y = \{0,1\}^n$. We also use the vector notation $\textbf{y} = (y_1, y_2, ..., y_n)$ to denote the observations of $Y$. To model the system, we can create a graph, or network, $\myg_1 = (\myv_1, \mye_1)$ where the node set $\myv_1 = \mathcal{A}$ and the edge set $\mye_1$ is the set of node pairs $(u, v) \in \myv_1 \times \myv_1 $ that are adjacent elements in at least one $S_i$. To represent $Y$ within the graph, we also create a sink node (i.e., a node with no out-edges) for each possible outcome.

In a HON $\myg_k = (\myv_k, \mye_k)$, $\myv_k \subseteq \mathcal{A}^k$ is a set of conditional nodes and $k > 1$ denotes the maximum order of the graph.  We denote each $u' \in \myv_k$ as a tuple, i.e., $u' = (a_0, a_1, ..., a_m)$ where $m <= k$. We refer to $a_m$ as the \textbf{base node} and each other $a_j$ as \textbf{conditions}, and in practice use the notation $u' = a_m|a_1, ..., a_{m-1}$ to emphasize the fact that $u'$ encodes a conditional distribution. In other words, $k$ indicates the maximum number of conditions that can be encoded in a single node. Like $\mye_1$, the edge set $\mye_k \subseteq \myv_k \times \myv_k$ is a set of node pairs; however, because each $u' \in \myv_k$ represents multiple entities, each pair of nodes $(u', v') \in \mye_k$ naturally represents a higher-order interaction while remaining a 2-tuple. Edges are directed such that $(u, v) \neq (v, u)$ and weighted via $w_k: \mye_k \xrightarrow[]{} \mathbb{R}_{\geq 0}$, determined by the population prevalence and where 0 indicates that there is no edge from $u$ to $v$.

\subsection{Detecting outcome-driven higher-order dependencies}
To address the key challenges presented above, we first define a measure of conditional entropy on a generic graph $\myg = (\myv, \mye)$ with respect to $Y$. This allows us to identify and encode the conditional nodes that we expect to reduce this entropy. First, we compute the class entropy $Y$ according to the standard formula:

\begin{equation}
    h(Y) = -\sum_{y \in Y} P(y)log_{2}P(y).
\end{equation}

To compute class entropy with respect to $\myg$, let us first assume that for each of the possible outcomes of $Y$, a corresponding sink node exists in $\myg$ (though, for notational simplicity, we will refer to $\myv$ as though it does not include these sink nodes). Let $\pi_{\myg} : \mathcal{A} \xrightarrow[]{} Y$ denote a random walker that begins at a random first-order node and continues traversing random edges in $\myg$ until it reaches one of the sink nodes. We define the conditional entropy of $Y$ given $\myg$ as follows:

\begin{equation}
    \label{eq:hyg}
    h(Y | \myg) = - \sum_{u \in \mathcal{A}} P(u) \sum_{y \in Y} P(\pi_{\myg}(u)=y)log_{2}P(\pi_{\myg}(u)=y),
\end{equation}
where $P(u)= \frac{\sum_{v \in \mathcal{N}(u)}w(u,v)}{\sum_{(u,v) \in \mye}w(u,v)}$ (i.e., the normalized, weighted out-degree of $u$) is the probability that the random walker starts at $u$. We can easily estimate $P(\pi_{\myg}(u)=y)$ by sampling random walks on $\myg$. Finally, we can formulate the information gain of $\myg$ with respect to $Y$ as:

\begin{equation}
    \label{eq:igyg}
    IG(Y, \myg) = H(Y) - H(Y|\myg),
\end{equation}

While the combinatorial nature of $\myg$ means that we cannot optimize $IG(Y, \myg)$ directly, we can do so indirectly by creating conditional nodes that will reduce the entropy of $\pi_{\myg}$. Since $\myg$ is a summary of the observed paths $\mathcal{S}$, nodes that provide meaningful information about $Y$ with respect to $\mathcal{S}$ will also reduce $IG(Y, \myg)$. Given a candidate conditional node $u' \in \mathcal{A}^k$, we can compute its conditional entropy according to:
\begin{equation}
    \label{eq:hyuprime}
    h(Y|u') = \sum_{y \in Y} P(y,u')log_{2}\frac{P(y,u')}{P(u')},
\end{equation}
where $P(u')$ is the probability that $u'$ is a \textbf{subsequence} of some $S_i$, i.e., that each $a_i \in u'$ is in $S_i$ and in the same relative order. Finally, we compute the information gain (IG) of $u'$ with respect to $Y$ as:
\begin{equation} 
    \label{eq:iguprime}
    IG(Y,u') = h(Y) - h(Y|u'),
\end{equation}

This formulation allows us to compute information gain for each candidate $u'$, but does not tell us how to interpret the computed value. Unsupervised HON methods use a user-defined threshold function, but this approach is ad hoc \cite{saebi2020efficient}. To more objectively determine if the $IG(Y,u')$ value for a candidate conditional node occurs due simply noise, rather than a true dependency, we can utilize a Fisher-Yates shuffle. To do so, we build a reference distribution for each candidate $u'$ by randomly permuting the outcome vector (\textbf{y}) 100 times and recomputing the information gain. We then directly compute the likelihood that the observed $IG(Y,u')$ is greater than expected from a distribution where the outcome has been randomly assigned. Using a $t$-statistic for $IG(Y,u')$, which we denote as $t_{u'}$, we construct the set of nodes as follows:


\begin{equation}
    \label{eq:vk}
    \myv_k = \mathcal{A} \ \bigcup \ \{u' \in \mathcal{A}^{k} :  t_{u'} \geq \alpha \},
\end{equation}
where $\alpha$ denotes the standard score threshold, which we default to 1.0.

\subsection{Skipping noisy data} After constructing $\myv_k$, we must determine which nodes to connect; i.e., how to construct $\mye_k$. Although prior HON methods construct edges in a strictly linear fashion \cite{saebi2020efficient,krieg2020growhon}, the supervised nature of our proposed approach allows us to focus the growth of the network to consider subsequences rather than substrings (i.e., where each $a_i \in u'$ is directly adjacent in $S_i$). This is demonstrated in our formulation of Eq. \ref{eq:hyuprime}, where we compute $IG(Y|u')$ in a way that does not enforce a strict adjacency constraint and enables the model to ``skip'' noisy diagnoses. We construct $\mye_k$ similarly. Specifically, for each $S_i$ in $\mathcal{S}$, we add edges such that each node $u' \in \myv_k$ that is also a subsequence of $S_i$ is connected to every node that succeeds it in $S_i$, including the outcome (sink) nodes. Although this approach removes the linear constraint and thus allows for much greater model expression, it has the drawback of increased computational cost, since each $S_i$ has $\sum_{j=1}^k{\binom{|S_i|}{j}}$ subsequences to consider. If we did not allow for skip steps, we would only need to consider $\sum_{j=1}^{k}{(|S_i|-j)}$ subsequences for each $S_i$. Some form of pre-pruning would mitigate this problem, but risks missing useful higher-order combinations. Fortunately, computing Eq. \ref{eq:vk} is embarrassingly parallelizable. Additionally, for larger data sets, computational overhead could be reduced by considering a minimum frequency threshold $x$ such that each $u'$ is only tested if it appears at least $x$ times in $\mathcal{S}$. Future work could explore additional strategies for increasing the sparsity of these skip connections.

\section{Experiments}

\subsection{Data}
For the experiments presented in this work, we leveraged a longitudinal claims dataset for type 2 diabetes (T2D) patients in the state of Indiana. The complexity of structural relationships between T2D comorbidities makes it a particularly compelling use case \cite{oh2016type}, and these particular data have been utilized in prior HON works to allow for direct comparison to existing baselines \cite{krieg2020higher}. Each $S_i$ represents a patient, and each entity in $S_i$ is an ICD9 major code representing that patient's diagnosis history. In total, the data comprises 908 distinct ICD9 codes across 913,475 distinct patients, all of whom received either a clinical diagnosis of T2D, reported a laboratory glycated hemoglobin (HbA1C) test result of at least 6.5\%, or was prescribed at least one Medi-Span-defined anti-diabetes medication. After preprocessing according to the procedure proposed by \cite{krieg2020higher} (discarding E and V codes, removing repeated diagnoses), each trajectory contained an average of 14.3 diagnoses.

Table \ref{tab:icd9ref} provides a summary of the diagnosis codes we frequently reference. Our experimentation centers on three patient outcomes that have been relevant to other studies of T2D trajectories: congestive heart failure (ICD9 code 428), hypotension (ICD9 code 458), and acute renal failure (ICD9 code 584) \cite{maule2007orthostatic,oh2016type,jeong2020temporal,krieg2020higher}. For each of these outcomes, we performed these additional preprocessing steps:
\begin{enumerate}
    \item We assigned a positive label to all patients for whom the outcome code appears in their trajectory, and a negative label to all others.
    \item For patients with a positive outcome, we removed both the outcome code and any subsequent diagnoses. This is because we are interested in predicting the outcomes, i.e. modeling the events that precede the outcome.
    \item We discarded any trajectories with fewer than 5 diagnoses preceding the outcome.
\end{enumerate}

Table \ref{tab:trajsum} summarizes the final set of trajectories for each outcome, which we then partitioned into training, validation, and testing sets according to an 80/10/10 ratio. In all cases, we built graphs and trained predictive models using only the training set. For predictive tasks, we used the validation set to select the best hyperparameters and report only the results as measured on the testing set. 

\begin{table}[t]
    \caption{Reference for ICD9 codes in the T2D data set.}
    \centering
    \begin{tabular}{ll}
    \toprule
    \textbf{Code} & \textbf{Description} \\ \midrule
    250 & Diabetes mellitus \\
    401 & Essential hypertension \\
    414 & Ischemic heart disease \\
    428 & Congestive heart failure \\
    458 & Hypotension \\
    461 & Sinusitis \\
    462 & Pharyngitis \\
    584 & Renal failure \\
    585 & Chronic kidney disease \\ \bottomrule
    \end{tabular}
    \label{tab:icd9ref}
\end{table}

\begin{table*}[ht]
\caption{Summary of outcomes for T2D patient trajectories.}
\label{tab:trajsum}
\centering
\begin{tabular}{lccccc}
\toprule
\textbf{Outcome} &\textbf{\# train/val/test} & \textbf{P(Y=1)} & \textbf{h(Y)} & $\mathbf{|\myv_1|}$ & $\mathbf{|\myv^*_4|}$ \\

\midrule
    Heart failure & 460,179/57,523/57,522 & 0.1068 & 0.4904 & 906 & 18,231 \\
    Hypotension & 473,736/59,218/59,217 & 0.0414 & 0.2485 & 906 & 45,554 \\
    Renal failure & 472,800/59,100/59,100 & 0.0480 & 0.2777 & 906 & 45,213 \\
\bottomrule
\end{tabular}   
\end{table*}

\subsection{Tasks and experimental setup}
To robustly assess the proposed model, we designed a series of analyses that quantify how much information each graph provides about an outcome and demonstrate their utility for traditional outcome-prediction tasks. For the proposed HON, we set $k=4$, which is both the highest value with which we could compute $\myg^*$ efficiently and the point at which higher-order dependencies saturated in prior work on HONs of T2D disease trajectories \cite{krieg2020higher}, and constructed $\myg^*_k$ for each of the aforementioned outcomes. We evaluated multiple graphs as baselines, including a FON ($\myg_1$), existing unsupervised HON ($\myg_4$) \cite{krieg2020growhon}, and ablated versions of the proposed HON without the skip-step ($\myg'_4$).

\subsubsection{Information Gain of Graphs and Paths} \label{sec:resultsinfogain} We first evaluated how well the global structure of each graph corresponded to meaningful disease pathways. To do this, we quantified the overall information of each graph with respect to each outcome using our formulation of $IG(Y, \myg)$ (Eq. \ref{eq:igyg}). Specifically, for each first-order node, we used a random walker to sample 10,000 paths. For each path, the walker began at the corresponding first-order node and continued traversing the graph until reaching one of the outcome (sink) nodes. We used these paths to compute $h(Y | \myg)$ (Eq. \ref{eq:hyg}), and subsequently $IG(Y, \myg)$. We repeated this procedure 10 times for each graph and reported the average and standard deviation. Metrics produced in this way are meaningful as they quantify the amount of information each graph provides as related to the purity of paths in reaching an outcome.

Next, we analyzed the local structures of our proposed model, $\myg^*_4$, by extracting the most informative nodes and paths. Using the same random walking procedure, we computed $IG(Y,u')$ (Eq. \ref{eq:iguprime}) for all nodes (including conditional nodes) in each graph. To compute the information gain for a path, we calculated the mean $IG(Y,u')$ for each $u'$ in the path. This procedure allows us to validate whether the informative nodes and paths in $\myg'_4$ match known clinical relationships between diagnoses and outcomes. 

Finally, we compared both the existence and informativeness of paths leading to a positive outcome to assess if the proposed supervision and skip-step resulted in meaningful differences in learned network structure. Utilizing 1,000,000 randomly sampled paths leading to the heart failure outcome, again extracted from the random walker, we identified the most informative trajectory (defined by the multiplicative weight of $IG(Y,u')$, as calculated above, for each node $u'$ in the trajectory) beginning at each of the 907 ICD codes in the dataset. We then calculated: 1) the percentage of paths that existed exactly as sampled from $\myg^*_4$ in each baseline network, and 2) the percentage of paths for which the IG in $\myg^*_4$ was higher than that in the comparison network. For paths that could be not exactly reproduced in a comparison network, we recursively pruned conditions from the higher-order nodes and attempted to continue the path. For example, if an edge from 403 to 414|403 did not exist, we attempted to traverse the baseline network using the edge from 403 to 414 directly. If a path was still unable to be found, no comparison was made to prevent biased denominators. We note that as the multiplication of IG across a trajectory will be biased based on path length, we grouped paths based on their exact length (2 through 6) before performing sensitivity analysis for ``top'' paths.

\subsubsection{Outcome Prediction} Second, we explored the use of $\myg^*_4$ for a classification task. Although a HON provides increased interpretability of trajectories, it is generated at the population level and not directly applicable to the prediction of an individual's outcome. As such, this analysis sought to determine how trajectories learned by a HON can be integrated into a supervised classification framework as compared to state-of-the-art sequence models (i.e., BERT-based models). In this case, the sequence of diagnoses in each disease trajectory serves as input to a model, whose job is to map the trajectory to a binary outcome (e.g., 1 for heart disease, and 0 for no heart disease). Our baseline, BEHRT, achieves this by 1) mapping each token (entity) from the trajectory to an embedding vector, and 2) passing the token embeddings through a stack of transformer encoder blocks \cite{li2020behrt}. Following the procedure proposed by the seminal BERT model \cite{devlin2018bert}, BEHRT pre-trains the encoder blocks via token masking and a stack of decoder blocks. Slight variants of this approach have been proposed specifically for disease prediction \cite{rasmy2021med,meng2021bidirectional}, but differences among them are minor, so we consider BEHRT as a representative baseline. Our experiments used the public code and hyperparameters as published by \cite{li2020behrt}.

To classify trajectories using a graph $\myg$, we map the problem to a subgraph classification task. Given an input trajectory, we use a graph neural network (GNN) to compute an embedding vector for each node. When $k>1$, we map tokens to higher-order nodes whenever possible, prioritizing combinations that are temporally close in the trajectory. For example, given an input trajectory $\langle ..., 461, ..., 250, ..., 401, ... \rangle$, if node $401|250 \in \myv^*_4$, then we use it instead of node $401$ to compute the GNN embedding for token $401$. If $401|250 \notin \myv^*_4$ and node $401|461 \in \myv^*_4$, then we use $401|461$. Following BEHRT, we pass the GNN embeddings forward through a stack of transformer encoder blocks to compute the outputs. Essentially, this means that we replace BEHRT's basic, fully-connected embedding layer (step 1 above) with a GNN, which learns embeddings mediated by the given graph structure. We refer to this model as Transformer/$\myg$. Additionally, we considered a model Transformer/$\myg$+Embs, which uses a concatenation of the fully connected embedding layer and the GNN embeddings as input to the encoder block. Intuitively, this approach can be described as a modification of BEHRT that uses embeddings from $\myg$ to augment the token embeddings. To evaluate the graph-based models, we used a simple two-layer graph convolutional network (GCN) as the GNN. For all models, we used 128 as the token embedding size. This means that, to avoid incurring an unfair advantage due to using more parameters, for Transformer/$\myg$+Embs, the output dimensions for the GCN and fully connected embedding layer were 64 each. We trained all models using the Adam optimizer, with a learning rate of 0.0001 and a batch size of 32. We trained and tested each model 10 times, and reported the mean and standard deviation of the area under the precision-recall curve (AUPRC), a standard measure of binary classification performance in class-imbalanced settings \cite{davis2006relationship,li2020behrt}.

\begin{table}[ht]
\caption{Summary of toy (pattern matching) problems.}
\label{tab:toyproblems}
\centering
\begin{tabular}{lcr}
\toprule 
\textbf{Name} & \textbf{Diagnosis Pattern} & \textbf{P(Y=1)} \\
\midrule
    Toy 1 & $\{461,462,463,464,465,466\} \xrightarrow[]{} 401 \xrightarrow[]{} 250$ & $0.0475$ \\
    Toy 2 & $401 \xrightarrow[]{} \{461,462,463,464,465,466\} \xrightarrow[]{} 250$ & $0.0644$ \\
    Toy 3 & $401 \xrightarrow[]{} 250 \xrightarrow[]{} \{461,462,463,464,465,466\}$ & $0.0736$ \\
\bottomrule
\end{tabular}
\end{table}

Ultimately, we extended the analysis of the classification task to directly assess performance on known trajectories with noisy labels. To accomplish this using the T2D trajectories, instead of labeling each trajectory according to a clinical outcome, we assigned labels according to whether the trajectory contained a specific pattern of diagnosis codes. For each of these patterns, which are summarized in Table \ref{tab:toyproblems}, we labeled each trajectory as positive (1) if it matched the pattern (while allowing for \textit{any number} of intermediate diagnoses), and negative (0) otherwise. The particular codes we chose to define these patterns were arbitrary and based on prevalence in the data set (i.e., to ensure that each pattern had a sufficient number of positive examples) rather than on known T2D comorbidities or outcomes. This classification problem is thus conceptually simple but contains the kinds of higher-order dependencies that are prevalent in real-world systems yet non-trivial for many sequence models to represent \cite{xu2016representing}. We also conducted experiments in which the trajectories were subject to various thresholds of label noise. Here, we denote as \textbf{noise ratio} the probability that a trajectory matching the pattern was randomly assigned a negative label. For example, at a noise ratio of $0.4$, only 60\% of the matching trajectories were assigned a positive label, while the remaining 40\% were assigned a (noisy) negative label. Label noise is ubiquitous in real-world applications---for example, in disease trajectory studies, a patient may appear to have a negative outcome only because the positive outcome has not yet been diagnosed and recorded---so it is essential to understand the effect of this type of noise on model performance \cite{frenay2013classification,han2020survey}.

\section{Results}
\label{sec:results}

\begin{table}[ht]
\caption{$IG(Y, \myg)$ for each graph and data set. Reported values are the means and standard deviations of 10 iterations of the random walking procedure. Bold font indicates the best result for each data set.}
\label{tab:infogains}
\centering
\begin{tabular}{lrrr}
\toprule 
& \textbf{Heart failure} & \textbf{Hypotension} & \textbf{Kidney failure} \\ \midrule
                     $\myg_1$ & $0.0001$ \scriptsize $\pm 0.003$ & $0.0037$ \scriptsize $\pm 0.002$ & $0.0044$ \scriptsize $\pm 0.002$ \\
      $\myg_4$ & $0.0097$ \scriptsize $\pm 0.003$  & $0.0138$ \scriptsize $\pm 0.001$ & $0.0159$ \scriptsize $\pm 0.003$ \\
      $\myg'_4$ & $0.0046$ \scriptsize $\pm 0.002$ & $0.0011$ \scriptsize $\pm 0.001$  & $0.0123$ \scriptsize $\pm 0.002$ \\
        $\myg^*_4$& $\mathbf{0.0892}$ \scriptsize $\pm 0.004$  & $\mathbf{0.0723}$ \scriptsize $\pm 0.001$  & $\mathbf{0.0836}$ \scriptsize $\pm 0.003$ \\
\bottomrule
\end{tabular}
\end{table}


\subsection{Information Gain of Graphs and Paths} \label{sec:igresults} Table \ref{tab:infogains} presents $IG(Y, \myg)$ for each graph and outcome. For all outcomes, we find that only those paths sampled from $\myg^*_4$ provided a substantial increase in the amount of information gain, indicating that the overall purity of the paths was highly distinct between those patients found to have the outcome and those who did not. Notably, despite being grown based on the relationship to the outcome, $\myg'_4$ did not perform overly well. This further suggests that without the ability to skip irrelevant diagnoses, even an outcome-driven HON model cannot detect and encode the most informative paths.

\begin{table}[ht]
\caption{Classification performance (AUPRC). Bold font indicates the best result for each data set.}
\label{tab:classification}
\centering
\begin{tabular}{lrrr}
\toprule 
\textbf{Model} & \textbf{Heart failure} & \textbf{Hypotension} & \textbf{Kidney failure} \\ \midrule
                   BEHRT      & $0.372$ \scriptsize $\pm 0.00$ & $0.133$ \scriptsize $\pm 0.00$ & $0.222$ \scriptsize $\pm 0.00$ \\
        BEHRT/$\myg_1$       & $0.305$ \scriptsize $\pm 0.02$ & $0.109$ \scriptsize $\pm 0.01$ & $0.172$ \scriptsize $\pm 0.01$ \\
        BEHRT/$\myg_1$+Embs  & $0.375$ \scriptsize $\pm 0.00$ & $0.134$ \scriptsize $\pm 0.00$ & $0.228$ \scriptsize $\pm 0.00$ \\
        BEHRT/$\myg^*_4$       & $0.321$ \scriptsize $\pm 0.01$ & $0.115$ \scriptsize $\pm 0.01$ & $0.190$ \scriptsize $\pm 0.01$ \\
        BEHRT/$\myg^*_4$+Embs  & $\mathbf{0.386}$ \scriptsize $\pm 0.01$ & $\mathbf{0.144}$ \scriptsize $\pm 0.01$ & $\mathbf{0.237}$ \scriptsize $\pm 0.01$ \\
\bottomrule
\end{tabular}
\end{table}

\begin{figure*}[ht]
    \centering
    \includegraphics[width=1.0\textwidth]{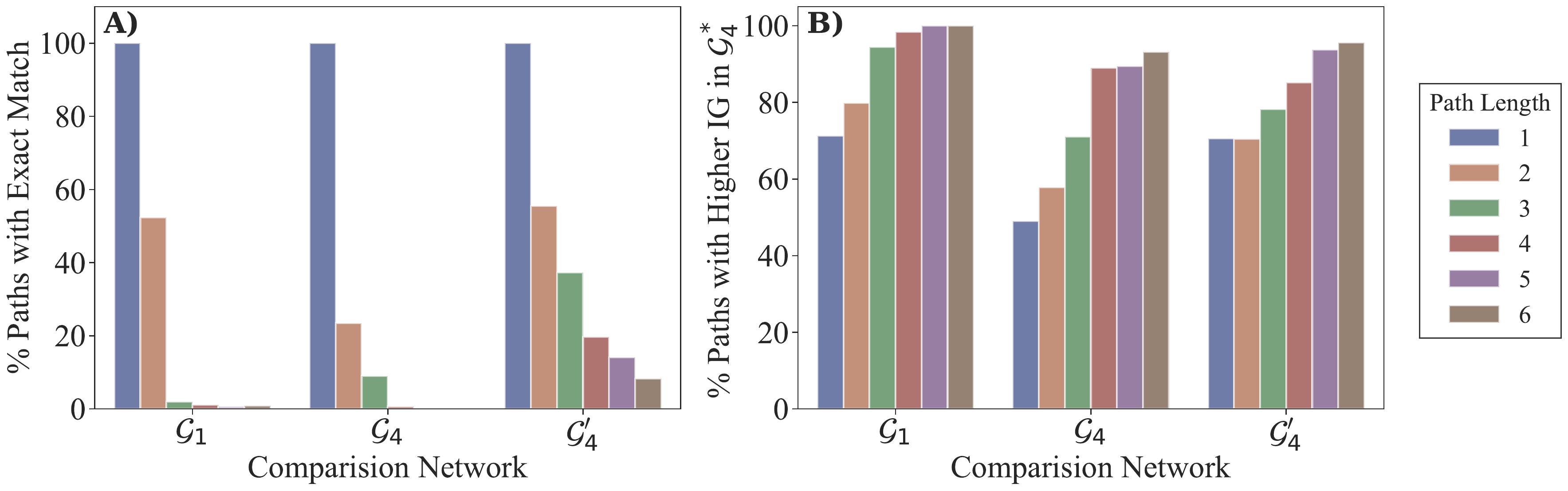}
    \caption{Comparison of paths towards heart failure (ICD9 code 428) sampled from $\myg_4$. Panel (A) shows the percentage of exact matches between paths of length 1-6, while panel (B) shows the percentage of paths for which the Information Gain (IG) was higher in $\myg_4$.} 
    \label{fig:validation}
\end{figure*}

Looking next to the comparison of sampled heart failure paths in $\myg^*_4$ with other graphs, Figure~\ref{fig:validation}A, demonstrates that, on average, well below 50\% of paths leading to heart failure in the proposed graph could be replicated in any of the comparison graphs. Note that a path of length 1 (i.e., a single node) will always have a 100\% match for all networks, as there is always a non-zero probability of reaching the outcome of interest from any given starting point. Unsurprisingly, shorter trajectories had a higher prevalence of overlap across the various network configurations. However, a small subset of the best paths for rare and/or unrelated root nodes were found to contain all first-order steps, allowing for some exact matches to the FON beyond a path length of 2. Critically, even after pruning higher-order dependencies at each step, we were only able to recover matches for just over 74\% of the paths in $\myg_1$, 62\% in $\myg'_4$, and only 40\% in $\myg_4$ across all sensitivity analyses, indicating that edges learned by $\myg^*_4$ never received enough support to even be grown at a first-order in these earlier methods. Additionally, the low match rate for the unsupervised HON $\myg_4$, even compared to the FON $\myg_1$, demonstrates how the dependencies detected by an unsupervised model cause the network to overfit on fragmented trajectories instead of encoding meaningful paths.

Moving beyond the simple existence of paths, Figure~\ref{fig:validation}B, illustrates that the individual paths in $\myg^*_4$ as compared to those extracted from baseline networks (evaluating exact matches when possible and, alternatively, pruned matches), as denoted by the average purity of the path in reaching a given outcome. Again, shorter paths have more similar performance across multiple networks, because the statistical support for paths that only include 1 or 2 nodes before an outcome is likely to be also found in the baseline networks. However, for paths with a length of at least 3, $\myg^*_4$ can generate paths with superior purity in over 90\% of cases. 

\subsection{Outcome Prediction} \label{sec:predresults} As Table \ref{tab:classification} shows, on the classification task, we observed consistent performance trends across all outcomes. First, we note that the transformer with only GNN embeddings (Transformer/$\myg$) significantly underperformed relative to BEHRT using either $\myg_1$ or $\myg^*_4$. On the other hand, augmenting the inputs with graph embeddings consistently improved transformer performance, especially with $\myg^*_4$. With a more expressive GNN implementation, these improvements would likely be even more substantial \cite{krieg2022deep}.

\begin{figure*}[ht]
    \centering
    \includegraphics[width=1.0\textwidth]{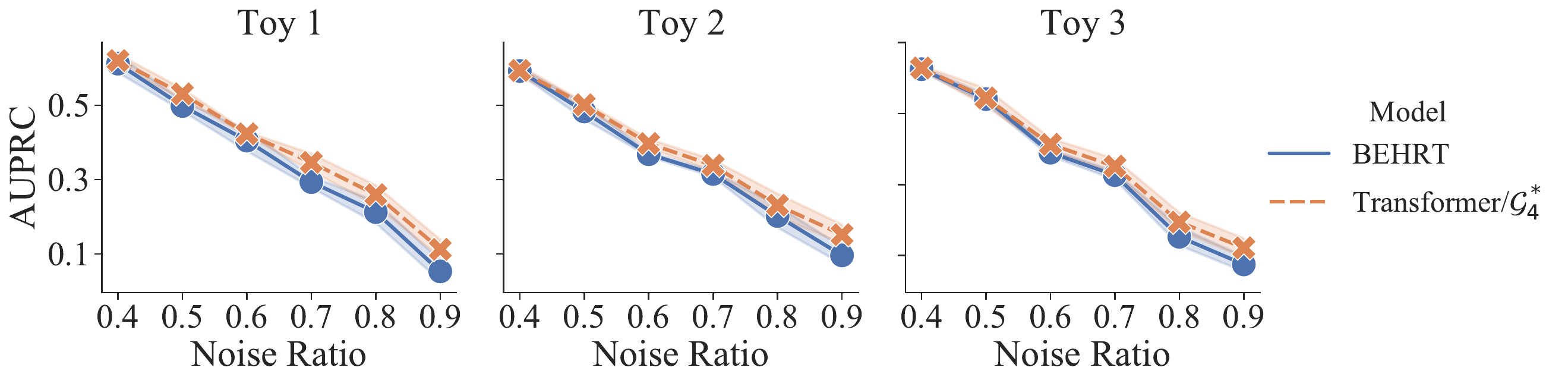}
    \caption{Classification results on toy trajectories at various levels of label noise. Shading represents standard deviation of 10 runs for each.}
    \label{fig:toypreds}
\end{figure*}

Figure \ref{fig:toypreds} summarizes the results of the classification experiments on the toy pattern matching task, which illuminate a key reason for the effectiveness of $\myg^*_4$. At a noise ratio of $0$, the task was trivial due to the ability of the transformer to match the higher-order patterns, and both models achieved AUPRC $\geq$0.999. While the noise ratio remained below 0.5, the performance of both models was virtually identical, so these are not shown to preserve scale. As the noise ratio increased, AUPRC consistently decreased for both models. While this was expected, we also noticed that a performance gap emerged as the noise ratio surpassed $0.5$. At the noisiest ratio of 0.9, the differences in AUPRC were relatively large in favor of Transformer/$\myg^*_3$+Embs over BEHRT: 0.112 and 0.053, respectively, on Toy 1 (+113\%); 0.152 and 0.096, respectively, on Toy 2 (+58\%), and 0.122 and 0.075, respectively, on Toy 3 (+62\%). These results suggest that at least one of the reasons that Transformer/$\myg^*_3$+Embs performs well on the outcome prediction tasks is that the embeddings from $\myg^*_3$ help improve the robustness of the encoder against label noise.

\section{Discussion}
\label{sec:rq4}
This work has introduced a novel implementation of higher-order networks able to extract meaningful trajectories towards an outcome of interest even in the presence of noise commonly found in real-world data. As evidenced by the results above, simply replacing the objective function for a HON model to enforce the growth of paths that maximize purity in reaching an outcome is not sufficient for detecting the most informative paths. This is hardly surprising in a healthcare context; disease trajectories encompass a variety of diagnoses elicited from various encounters within the health system, many of which are unlikely to be relevant to the outcome of concern. In this case, these noisy intermediate diagnoses only function to reduce the statistical support for the relevant combinations of diagnoses, making it less likely that a model can identify them.

We see strong evidence of this phenomenon in the 
\textit{Information Gain} analysis (Section \ref{sec:igresults}). As expected, the supervised HON without a skip step ($\myg'_4$) had the highest percent of matching paths to our proposed model, as both are designed to grow paths and create higher-order dependencies using the same objective function. However, less than 20\% of paths $\geq$4 matched exactly without the inclusion of the skip-step, strongly illustrating how noise from unrelated diagnoses can fragment trajectory growth. Looking deeper, we note that 60\% of the paths extracted from the proposed network could be reconstructed in this non-skip step network by pruning higher-order edges; i.e. edges between relevant diagnoses were created, but conditional higher-order paths were not constructed given noise. As it is clear these reduced paths are not as informative (Figure~\ref{fig:validation}B), investigation of higher-order nodes can provide additional insight into what information is lost. 

In addition to analyzing paths in $\myg'_4$, we can also study individual nodes. As Table \ref{tab:trajsum} shows, for the heart failure, hypotension, and renal failure outcomes, the proposed method created 17,325, 44,648, and 44,307 conditional nodes, respectively, in each $\myg^*_4$. In the case of heart failure, 27\% of these were second-order nodes, 62\% were third-order nodes, and the remaining 11\% were fourth-order nodes. The five most common conditions were chronic ischemic heart disease (ICD9 code 414), essential hypertension (401), hypercholesterolemia (272), cardiac dysrhythmia (427), and anemia (285)---all of which are known to be strongly correlated with heart failure\cite{van2014co,groenewegen2020epidemiology}. Of the 17,325 conditional nodes, 8,372 (48\%) were conditioned one at least one of these five diagnoses. We observed similar patterns by analyzing the individual nodes with the highest IGs. Almost all these nodes used a base diagnosis of cardiomyopathy (IC9 code 425) in conjunction with various conditions; for example, 425$|$511$|$427 (cardiomyopathy given pleurisy given cardiac dysrhythmia) produced the highest IG (0.051). Besides validating that the discovered nodes are truly informative of the heart disease outcome, examining these nodes highlights certain combinations of risk factors and drive further investigation. 




Turning next to the utilization of the identified trajectories, the \textit{Outcome Prediction} analysis (Section \ref{sec:predresults}) provided a novel approach to improve the interpretability of sequence prediction tasks while maintaining predictive performance. The interplay between these factors is particularly relevant in health-centric applications, where the utility and adoption of computational models rely not only on a measure of accuracy but increasingly on an ability to investigate the factors underlying output, which may be used to guide clinical analyses and recommendations~\cite{rasheed2022explainable,amann2020explainability,yang2022explainable}. 

Given the established state-of-the-art predictive performance and widespread availability in software packages, deep learning transformer architectures are unlikely to be replaced as a standard model for sequence prediction tasks. Yet, it is clear there is no straightforward way for BEHRT or similar models to isolate the trajectories that drive its outputs. For example, while we can inspect the attention weights to identify that a diagnosis code $414$ has a probability of transitioning to $250$, and that $250$ then has some probability of reaching a positive outcome, this transitive interpretation cannot account for the complex dependencies learned by the model over all \textit{combinations} of diagnoses and may be misleading \cite{chefer2021transformer}. As a result, we sought to augment these models rather than replace them entirely.

By utilizing a graph to augment the embeddings used for prediction, we can easily and directly examine the relationships between learned edges to compute, for example, the probability distribution over $\pi_{\myg}$ (Eq. \ref{eq:hyg}). We can use this distribution to understand how different risk factors could contribute to a prediction. We know by our definition of Eq. \ref{eq:vk} that grown edges must provide some information about the probability of reaching the outcome. In the context of a higher-order network, the fact that each node can represent the conditional probability across multiple entities helps us then to understand how these combinations of comorbidities can affect outcome probabilities in a way that transcends the previously-discussed limitations of transformers~\cite{krieg2020growhon}. For example, the node 425$|$511$|$427 produced the highest IG with respect to heart failure in $\myg'_4$ (0.051). On their own, we observed IGs of 0.022 and 0.004 for cardiomyopathy (425) and pleurisy (511), respectively, both ranking within the top 15 most informative first-order nodes. For cardiac dysrhythmia (427), on the other hand, we only observed an IG of 0.001---rank 76 of the 906 first-order nodes. And while we saw an increase in IG for the second-order nodes 425$|$511 (0.026) and 511$|$427 (0.010) when compared to their first-order counterparts 425 and 511, respectively, these values are still much smaller than the information provided by 425$|$511$|$427. The progression of a patient over this third-order sequence of diagnoses thus seems to increase their probability of reaching heart failure in such a way that is more than the sum of its first or second-order parts. This means that the ability to examine such higher-order combinations and probabilities directly is a critical piece of a truly interpretable model.

Further, we are often led to believe there must be a trade-off between accurate, complex models and those that are interpretable. As a result, a simple non-inferiority of the augmented network in terms of predictive performance would be valuable. However, as we examine the results of this augmentation, we find it promising that results not only equal but slightly exceed that of the existing transformer. Other studies have noted that a transformer's dense embedding layer is equivalent to a fully connected graph and characterized the learning process of a transformer as learning the edge weights of an unspecified graph \cite{velivckovic2023everything}. Therefore, it is not surprising to see that such a model would overfit on noisy patterns, where the additional information gained by the more targeted graph provides useful information in learning weights for the transformer model. This idea is supported by the results of our analysis of the toy paths. We see a clear pattern in which the augmented graph+transformer model improves performance as the proportion of noise injected into a path increases. Such noise is abundant in healthcare data because of causes like time (e.g., a T2D patient has not \textit{yet} been diagnosed with heart disease) and false negatives, so robustness to it is an extremely useful characteristic of any model.

\bmhead{Limitations}
There are several limitations to the work presented here. First, although a large dataset was used to assess the performance of the proposed methodology on three distinct disorders and three toy examples, we recognize the data used only claims from a single state. Second, this model evaluated only trajectories of diagnoses. It is unlikely diagnosis history is the only factor driving a patient to reach a specific clinical outcome. As such, performance may be improved with the inclusion of a more comprehensive set of clinical factors (procedures, laboratory results, etc.). Current work is underway by the study team to extend the methodology to account for sequences of heterogeneous nodes. Finally, it is becoming increasingly important to consider the computational cost of the models we build. In some cases, the marginal improvements of the proposed model on the classification performance (Table \ref{tab:classification}) may not be worth the additional computation cost of constructing $\myg^*_3$ and using a GNN.

\section{Conclusion}
Given the way healthcare data are captured and utilized, noise is the norm rather than the exception. Drawing on the strengths of state-of-the-art work in EHR data analysis, disease trajectory modeling, and higher-order networks, this manuscript proposed a new method for constructing a HON in a supervised learning context. Different from prior approaches, the proposed method identifies the higher-order dependencies that are most informative of an outcome while skipping noisy or irrelevant steps. Experiments demonstrated that HONs constructed with the proposed method can provide significantly more information about an outcome than other graphs, and can successfully detect and encode dependencies that correspond to known comorbidities and disease pathways in T2D patients. Additionally, the proposed method's robustness to noise can be used to augment transformer-based models, marginally improving their performance on a disease prediction task when label noise is present. These results suggest several fruitful avenues for future work, such as improving the scalability of the proposed method to higher orders and longer trajectories, incorporating more granular information from trajectories (e.g., time intervals and heterogeneous/multimodal data) into the HON structure, and designing more sophisticated methods for using information from the HON to augment neural sequence models.

\bibliography{main}

\end{document}